\documentclass{article}

\usepackage{arxiv}

\usepackage[utf8]{inputenc} 
\usepackage[T1]{fontenc}    
\usepackage{hyperref}       
\usepackage{url}            
\usepackage{booktabs}       
\usepackage{amsfonts}       
\usepackage{nicefrac}       
\usepackage{microtype}      
\usepackage{lipsum}
\usepackage{verbatim}
\usepackage{array}
\usepackage{multirow}
\usepackage{graphicx}
\usepackage{amssymb,amsmath}
\usepackage{pgf}
\usepackage{tikz}
\usepackage{hyperref}
\usepackage[numbers]{natbib}
\usetikzlibrary{arrows,shapes,snakes,automata,backgrounds,fit,petri,shapes.multipart,calc,positioning,shapes.geometric}

\newcommand{\Mcal}{\mathcal{M}}%

\newcommand{\ptype}{\mathit{type}}
\DeclareMathOperator*{\argmax}{arg\,max}

\renewcommand{\citet}[1]
{\citeauthor{#1}}

\newif\ifshortversion
\shortversionfalse
\ifshortversion
\newcommand{\cutout}[1]{}
\newcommand{\shortened}[1]{#1}
\else
\newcommand{\cutout}[1]{#1}
\newcommand{\shortened}[1]{}
\fi

\title{Adapting a Kidney Exchange Algorithm to Align with Human Values}

\author{
  Rachel Freedman\thanks{Corresponding author. Email: \texttt{rachel.freedman@berkeley.edu}.} \\
  University of California, Berkeley\\
  \And
  Jana Schaich Borg \\
  Duke University\\
  \And
  Walter Sinnott-Armstrong \\
  Duke University \\
  \And 
  John P. Dickerson \\
  University of Maryland \\
  \And
  Vincent Conitzer \\
  Duke University
}

\graphicspath{ {fig/} }

\begin{document}
\maketitle

\begin{abstract}
    The efficient and fair allocation of limited resources is a classical problem in economics and computer science.  In kidney exchanges, a central market maker allocates living kidney donors to patients in need of an organ.  Patients and donors in kidney exchanges are prioritized using ad-hoc weights decided on by committee and then fed into an allocation algorithm that determines who gets what---and who does not.  In this paper, we provide an end-to-end methodology for estimating weights of individual participant profiles in a kidney exchange.  We first elicit from human subjects a list of patient attributes they consider acceptable for the purpose of prioritizing patients (e.g., medical characteristics, lifestyle choices, and so on).  Then, we ask subjects comparison queries between patient profiles and estimate weights in a principled way from their responses.  We show how to use these weights in kidney exchange market clearing algorithms.  We then evaluate the impact of the weights in simulations and find that the precise numerical values of the weights we computed matter little, other than the ordering of profiles that they imply.  However, compared to not prioritizing patients at all, there is a significant effect, with certain classes of patients being (de)prioritized based on the human-elicited value judgments. 
\end{abstract}

\section{Introduction}\label{sec:intro}

    As AI is deployed increasingly broadly, AI researchers are confronted with the moral implications of their work. The pursuit of simple objectives, such as minimizing error rates, maximizing resource efficiency, or decreasing response times, often results in systems that have unintended consequences when they confront the real world, such as discriminating against certain groups of people~\cite{ONeil17:Weapons}. It would be helpful for AI researchers and practitioners to have a general set of principles with which to approach these  problems~\cite{Wallach08:Moral,Tolchinsky12:Deliberation,Greene16:Embedding,Conitzer17:Moral,Noothigattu18:Voting}.
    
    \cutout{
    One may ask why any moral decisions should be left to computers at all.  There are multiple possible reasons.  One is that the decision needs to be made so quickly that calling in a human for the decision is not feasible, as would be the case for a self-driving car having to make a split-second decision about whom to hit~\cite{Bonnefon16:Social}.
    Another reason could be that each individual decision by itself is too insignificant to bother a human, even though all the decisions combined may be highly significant morally---for example, if we were to consider the moral impact of each advertisement shown online.  A third reason is that the moral decision is hard to decouple from a computational problem that apparently exceeds human capabilities.  This is the case in many machine learning applications (e.g., should this person be released on bail? \cite{Kleinberg2017:Human}), but also in other optimization problems.  
    }
    
    We are interested in\cutout{ one such problem:} the clearing house problem in {\em kidney exchanges}.   In
    a kidney exchange, patients who need a kidney transplant and have a willing
    but incompatible live donor may attempt to trade their donors' kidneys~\cite{Roth04:Kidney}.  Once these
    people appear at an exchange, we face a highly complex problem of deciding
    who matches with whom.  In some exchanges, this matching problem is solved
    using algorithms developed in the AI community: the United States~\cite{Dickerson15:FutureMatch}, the United Kingdom~\cite{Manlove15:Paired}, the Netherlands~\cite{Glorie14:Kidney}, and so on~\cite{Biro17:Kidney}.
    
    In this paper, we investigate the following issue.  Suppose, in principle, that we prioritize certain patients over others---for example, younger patients over older patients.  To do so would clearly be a morally laden decision.  How should this affect the role of the AI researcher developing these systems?  From a purely algorithmic perspective, it may seem that there is little more to this than to change some weights in the objective function accordingly.  But we argue that our job, as AI researchers, does not end with this simple observation.  Rather, we should be closely involved with the process for determining these weights, both because we can contribute  technical insights that are useful for this process itself, and because it is our responsibility to understand the consequences to which these weights will lead. The methodology that we develop integrates this prioritization into our development work.

\subsection{%
Our Contributions}
In this paper, we provide an end-to-end methodology for estimating weights of individual patient profiles in a kidney exchange, where these weights are used only for tiebreaking purposes (i.e., when multiple solutions give the maximal number of transplants).  

Executing our methodology in such a way that we would advocate directly adopting the results in practice would require substantially more effort and participation from other parties. For example, we would need to consult domain experts to determine which patient characteristics should be used to determine edge weights. We would also need to involve stakeholders such as policy-makers, doctors, and kidney exchange participants in the process for determining weights. For this reason, we execute this methodology in a limited fashion as a proof-of-concept, and evaluate the results in simulations.

We first elicit from human subjects a list of patient attributes they consider acceptable for the purpose of prioritizing patients in kidney exchanges (e.g., most subjects did not find race an acceptable attribute for prioritization).  Then, we ask subjects comparison queries between patient profiles that differ only on acceptable attributes, and estimate weights from their responses.  We show how to use these weights in kidney exchange market clearing algorithms, to break ties among multiple maximum-sized solutions.  We then evaluate the impact of the weights in simulations.
We find that the precise numerical values of the weights we computed matter little, other than the ordering of profiles that they imply.  However, compared to not prioritizing patients at all, there is a significant effect.  Specifically, the difference is experienced by donor-patient pairs that have an ``underdemanded''~\cite{Ashlagi14:Free,Toulis15:Design} combination of blood types; for them, their chances rise or drop significantly depending on their tiebreaking weights.

\section{Kidney Exchange Model}\label{sec:prelims}
We briefly review the standard mathematical model for kidney exchange and techniques from the AI community used to clear real kidney exchanges, and then give illustrative examples where tiebreaking would or would not play a role.

\subsection{Graph Formulation}\label{sec:prelims-graph}
In this work, as is standard~\cite{Roth04:Kidney,Roth05:Kidney,Roth05:Pairwise}, we encode an instance of a kidney exchange as a directed \emph{compatibility graph} $G = (V,E)$.  We first construct one vertex for each patient-donor pair in the pool.  Then, we construct an edge $e$ from vertex $v_i$ to vertex $v_j$ if the patient in $v_j$ wants and is compatible with the donor kidney of $v_i$.  A paired donor is willing to give her kidney if and only if the patient in her vertex $v_i$ receives a kidney.

Most fielded exchanges also assign a weight $w_{e}$ to an edge $e$. The function determining the weight for an edge is often opaque and set in an ad-hoc fashion by a committee. For example, a recent report \cite{UNOS15:Revising} proposes revising the policy for setting edge weights to incorporate the patient's ``calculated reactive panel antibody'' (which influences their likelihood of finding a match) and whether the pair has previously been a part of a ``failed exchange'' (in which the donor donates a kidney, but their corresponding patient does not receive one). The policy already includes a multitude of other factors, including which hospital registered the pair and whether the patient has previously donated an organ.\footnote{For a more detailed look into the inner workings of this process that sets edge weights, we direct the reader to a recent report by the UNOS US-wide kidney exchange~\cite{UNOS15:Revising}.} 

\begin{figure}[ht!bp]
\centering

\makeatletter
\tikzset{circle split part fill/.style  args={#1,#2}{%
 alias=tmp@name, 
  postaction={%
    insert path={
     \pgfextra{%
     \pgfpointdiff{\pgfpointanchor{\pgf@node@name}{center}}%
                  {\pgfpointanchor{\pgf@node@name}{east}}%
     \pgfmathsetmacro\insiderad{\pgf@x}
      \fill[#1] (\pgf@node@name.base) ([xshift=-\pgflinewidth]\pgf@node@name.east) arc
                          (0:180:\insiderad-\pgflinewidth)--cycle;
      \fill[#2] (\pgf@node@name.base) ([xshift=\pgflinewidth]\pgf@node@name.west)  arc
                           (180:360:\insiderad-\pgflinewidth)--cycle;            
         }}}}}  
 \makeatother  

\tikzstyle{altruist}=[circle,
  thick,
  minimum size=1.2cm,
  draw=black!50!green!80,
  fill=black!20!green!20
]

\begin{tikzpicture}[>=latex]
  
  \node (p_k) [shape=circle split,
    draw=gray!50,
    line width=1mm,text=black,font=\bfseries,
    circle split part fill={blue!20,red!20},
    minimum size=1.5cm,
  ] at (6,5) {$d_3$ (\texttt{A})\nodepart{lower}$p_3$ (\texttt{B})};

  \node (p_i) [shape=circle split,
    draw=gray!50,
    line width=1mm,text=black,font=\bfseries,
    circle split part fill={blue!20,red!20},
    minimum size=1.5cm,
  ] at (0,5) {$d_1$ (\texttt{A})\nodepart{lower}$p_1$ (\texttt{B})};

  \node (p_j) [shape=circle split,
    draw=gray!50,
    line width=1mm,text=black,font=\bfseries,
    circle split part fill={blue!20,red!20},
    minimum size=1.5cm,
  ] at (3,5) {$d_2$ (\texttt{B})\nodepart{lower}$p_2$ (\texttt{A})};

  \draw[->] (p_i) edge [bend left=20] (p_j);
  \draw[->] (p_j) edge [bend left=20] (p_i);
  \draw[->] (p_j) edge [bend left=20] (p_k);
  \path[->] (p_k) edge [bend left=20] (p_j);

\end{tikzpicture} 
\caption{A compatibility graph with three patient-donor pairs and two possible $2$-cycles.  Donor and patient blood types are given in parantheses.}\label{fig:simpletie}
\end{figure}
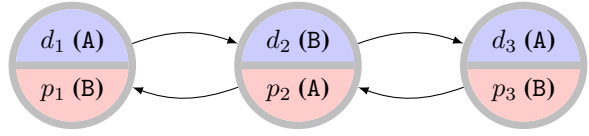

The weight $w_{e}$ of the edge $e$ from vertex $v_i$ to vertex $v_j$ roughly represents the utility to $v_j$ of obtaining $v_i$'s donor kidney, but can also be used to (de)prioritize specific classes of patient or donor, as we discuss later.  A cycle $c$ represents a possible sequence of transplants, with each vertex in $c$ obtaining the kidney of the previous vertex.  We use the term \emph{$k$-cycle} to refer to a cycle with exactly $k$ pairs.  For example, the compatibility graph in Figure~\ref{fig:simpletie} includes two possible $2$-cycles: a $2$-cycle between vertex $v_1$ and $v_2$, and a different $2$-cycle between vertex $v_2$ and $v_3$.  In kidney exchange, cycles of length at most some small constant $L$ (typically, $L \in \{2, 3, 4\}$) are allowed---all transplants in a cycle must be performed simultaneously so that no donor backs out after his patient has received a kidney but before he has donated his kidney.

Many fielded kidney exchanges gain great utility through the use of \emph{chains}~\cite{Montgomery06:Domino,Rees09:Nonsimultaneous,Anderson15:Finding,Ashlagi17:Need}.  Chains start with an altruist donor donating her kidney to a patient, whose paired donor donates his kidney to another patient, and so on.  In the standard model, altruistic donors are represented in the same way as patient-donor pairs, but with so-called ``dummy'' patients who are compatible with every patient-donor pair, yet do not require a kidney.  In this way, altruists and patient-donor pairs---as well as cycles and chains---can be treated similarly in optimization models.

A \emph{matching} $M$ is a set of disjoint cycles and chains in the compatibility graph $G$.  
There can be length limits on these cycles and chains, as discussed above, resulting in a smaller set of {\em legal matchings}.
The cycles and chains must be disjoint because no donor can give more than one of her kidneys (some recent work explores multi-donor donation~\cite{Ergin17:Multi,Farina17:Operation} but we do not consider this here).  Given the set of all legal matchings $\Mcal{}$, the \emph{clearing house problem} is to find a matching $M^*$ that maximizes utility function $u : \Mcal{} \to \mathbb{R}$.  Formally:

\vspace{-.4cm}
$$
M^* \in \argmax_{M \in \Mcal{}} u(M)
$$

Kidney exchanges typically use a \emph{utilitarian} utility function that finds the maximum weighted cycle cover (i.e., $u(M) = \sum_{c \in M} \sum_{e \in c} w_{e}$).  This can favor certain classes of patient-donor pairs while marginalizing others, a behavior we investigate later in this paper in the context of setting specific edge weights.  Alternate utility functions can be used to enforce incentive properties via mechanism design~\cite{Ashlagi14:Free,Li14:Egalitarian,Hajaj15:Strategy-Proof,Blum17:Opting,Mattei17:Mechanisms}.

\subsection{Clearing Kidney Exchanges}\label{sec:prelims-ip}
We briefly discuss optimization methods for clearing kidney exchanges; later, we show how to augment these methods to incorporate the ideas in this paper.  The standard clearing house problem for finite cycle cap $L>2$ (even without chains) is NP-hard~\cite{Abraham07:Clearing,Biro09:Maximum}, and is also hard to approximate~\cite{Biro07:Inapproximability,Luo16:Approximation,Jia17:Efficient}.  Thus, fielded kidney exchanges use integer program (IP) formulations to solve this difficult combinatorial optimization problem.

The first approach to clearing large kidney exchanges, due to~\citet{Abraham07:Clearing}, built a custom branch and price~\cite{Barnhart98:Branch-and-Price} integer program solver; generalizations of, and improvements on, their basic model have addressed scalability issues~\cite{Glorie14:Kidney,Anderson15:Finding,Dickerson16:Position,Dickerson18:Failure}.  We build a similar model in this work.

Formally, denote the set of all chains of length at most $K$ and cycles of length no greater than $L$ by $C(L,K)$.  Create a binary variable $x_c \in \{0,1\}$ for every $c \in C(L,K)$, and let $w_c = \sum_{e \in c} w_e$; then, solve the following integer program:

\vspace{-0.25cm}
\begin{equation*}
\max \sum_{c \in C(L,K)} w_c \ x_c \qquad
\mathit{s.t.} \qquad
\sum_{c : v \in c} x_c \leq 1 \quad \forall v \in V.
\end{equation*}
\vspace{-0.25cm}

The final matching is the set of \cutout{chains and cycles }$c$ such that $x_c = 1$.  
In this paper, we compare to a baseline where all edge weights are $1$, so that a maximum-cardinality solution is sought.  We then break ties in these solutions based on prioritization weights determined according to the procedure outlined in this paper.

\subsection{Tiebreaking and Prioritization: Examples}\label{sec:prelims-example}
Consider again the compatibility graph given in Figure~\ref{fig:simpletie}.  Here, there is one pair with a patient of blood type A and a donor of blood type B, and two pairs with a patient of blood type B and a donor of blood type A.  One of the latter two pairs will have to remain unmatched; either way, we obtain a solution of maximum cardinality (two vertices matched).  The standard algorithm may choose either solution; which one is chosen depends on details of the solver.
We may wish to break the tie based on other attributes of the two patients with blood type B, such as their age.  We will explore this in this paper.

\begin{figure}[ht!bp]
\centering
\makeatletter
\tikzset{circle split part fill/.style  args={#1,#2}{%
 alias=tmp@name, 
  postaction={%
    insert path={
     \pgfextra{%
     \pgfpointdiff{\pgfpointanchor{\pgf@node@name}{center}}%
                  {\pgfpointanchor{\pgf@node@name}{east}}%
     \pgfmathsetmacro\insiderad{\pgf@x}
      \fill[#1] (\pgf@node@name.base) ([xshift=-\pgflinewidth]\pgf@node@name.east) arc
                          (0:180:\insiderad-\pgflinewidth)--cycle;
      \fill[#2] (\pgf@node@name.base) ([xshift=\pgflinewidth]\pgf@node@name.west)  arc
                           (180:360:\insiderad-\pgflinewidth)--cycle;            
         }}}}}  
 \makeatother  

\tikzstyle{altruist}=[circle,
  thick,
  minimum size=1.2cm,
  draw=black!50!green!80,
  fill=black!20!green!20
]

\begin{tikzpicture}[>=latex]

  \node (p_i) [shape=circle split,
    draw=gray!50,
    line width=1mm,text=black,font=\bfseries,
    circle split part fill={blue!20,red!20},
    minimum size=1.8cm,
  ] at (-2,3) {$d_1$ (\texttt{AB})\nodepart{lower}$p_1$ (\texttt{O})};

  \node (p_j) [shape=circle split,
    draw=gray!50,
    line width=1mm,text=black,font=\bfseries,
    circle split part fill={blue!20,red!20},
    minimum size=1.8cm,
  ] at (0,5) {$d_2$ (\texttt{O})\nodepart{lower}$p_2$ (\texttt{AB})};
  
  \node (p_k) [shape=circle split,
    draw=gray!50,
    line width=1mm,text=black,font=\bfseries,
    circle split part fill={blue!20,red!20},
    minimum size=1.8cm,
  ] at (2,3) {$d_3$ (\texttt{AB})\nodepart{lower}$p_3$ (\texttt{A})};

  \node (p_l) [shape=circle split,
    draw=gray!50,
    line width=1mm,text=black,font=\bfseries,
    circle split part fill={blue!20,red!20},
    minimum size=1.8cm,
  ] at (4,5) {$d_4$ (\texttt{A})\nodepart{lower}$p_4$ (\texttt{O})};

  \draw[->] (p_i) edge [bend left=20] (p_j);
  \draw[->] (p_j) edge [bend left=20] (p_i);
  \draw[->] (p_j) edge [bend left=20] (p_k);
  \draw[->] (p_k) edge [bend left=20] (p_j);
  \draw[->] (p_j) -- (p_l);
  \draw[->] (p_l) -- (p_k);

\end{tikzpicture} 
\caption{A compatibility graph with four patient-donor pairs and two maximal solutions.  Donor and patient blood types are given in parentheses.}\label{fig:twomaximal}
\end{figure}
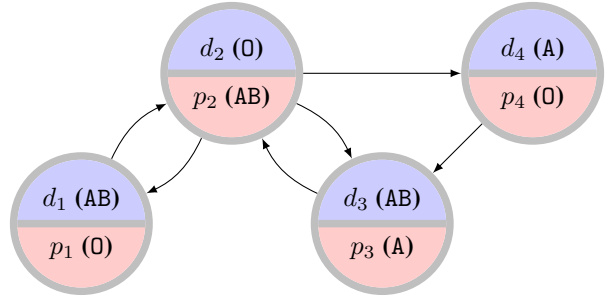

Now, consider the graph in Figure~\ref{fig:twomaximal}.  This graph has two maximal solutions. (A solution is maximal if it is not possible to include any other vertices without dropping others from the solution). One consists of the 3-cycle with vertices
AB-O, O-A, and A-AB (patient listed first in each case).  The other consists of the 2-cycle with vertices AB-O and O-AB. (For a complete description of which patient and donor blood types are compatible, see Table~\ref{tab:Compatibility}.)
The standard algorithm must choose the 3-cycle, because it matches more vertices.
While in principle one might consider choosing the 2-cycle, arguing that (due to other attributes) it is more important to save the patient from the O-AB vertex than it is to save {\em both} the patient from the O-A vertex {\em and} the patient from the A-AB vertex, in this paper we will not do so; we will always choose the 3-cycle, no matter what the values of the additional attributes are.

\begin{table}[h!]
\begin{center}
\begin{tabular}{llllll}
                       &    & \multicolumn{4}{c}{Patient} \\
                       &    & A     & B    & AB    & O    \\
\cline{3-6}
\multirow{4}{*}{Donor} & \multicolumn{1}{l|}{A}  & \checkmark     &      & \checkmark     &      \\
                       & \multicolumn{1}{l|}{B} &       & \checkmark    & \checkmark     &      \\
                       & \multicolumn{1}{l|}{AB} &       &      & \checkmark     &      \\
                       & \multicolumn{1}{l|}{O}  & \checkmark     & \checkmark    & \checkmark     & \checkmark   
\end{tabular}
    \vspace{10pt}
 \caption{\label{tab:Compatibility}Patient-donor blood-type compatibility. A checkmark denotes compatibility between the patient blood type in the column heading and the donor blood type in the row heading. For example, patients with blood type AB are compatible with all donor blood types, and donors with blood type O are compatible with all patient blood types.}
\end{center}
\end{table}

\section{Determining and Using  Prioritization Weights}\label{sec:moral}

In this section, we describe our procedure for computing prioritization weights and integrating them into the algorithm for clearing kidney exchanges. Because this procedure was intended as a proof-of-concept, we gathered preference data from participants recruited through the online platform Amazon Mechanical Turk (``MTurk'').\footnote{All experiments were conducted between fall 2016 and summer 2017.} However, if this procedure were used in a real-life kidney exchange, medical experts and other stakeholders would need to be involved in the process of determining weights. \footnote{That being said, it is not immediately clear what the optimal mix of stakeholders would be.  For example, it does not seem that medical training is especially helpful for evaluating how important it is whether a patient has dependents, such as small children.}

\subsection{Selecting Attributes}

First, we determined which patient attributes 
to include in our model by assessing which attributes a pool of human participants found acceptable to use for this purpose.  The attributes were generated by the participants in an open-ended survey to minimize experimenter bias. Specifically, participants ($N = 100$) were asked to read a brief description of the kidney transplant waiting list process, and then asked to imagine that a country is developing a new policy for allocating kidneys to patients on the waiting list.  Each participant was asked to report four potential patient attributes that they thought the kidney allocation policy ``morally ought to take into account,'' and four attributes that they thought the policy ``morally ought NOT to take into account.'' Each participant received \$0.85 as compensation for their participation. 

Participants' responses were independently sorted into attribute categories, including those listed in Table~\ref{tab:Attributes}, by two different researchers. Attributes that the UNOS algorithm already takes into account, such as patient-donor medical compatibility, were discarded.  The number of participants who mentioned each of the remaining attributes is noted in Table~\ref{tab:Attributes}. Because we are interested in improving the kidney allocation process, we only included those categories that more survey participants thought \textit{ought} to be taken into account than participants thought \textit{ought not} to be taken into account.

Because participants were asked to propose these attributes themselves, these results reflect which attributes occurred to them during the survey. This may skew the results in favor of attributes that seem more directly relevant to the medical context. For example, it's possible that 30 of the survey participants would have answered ``yes'' if directly asked whether criminal record should be taken into account, but because this aspect of personal life is not clearly related to health, only a few of those thought of it when prompted to consider public health policy during the survey. Additionally, we explicitly listed age as an example of the sort of attribute the policy might consider, which likely biased participants toward including it in their responses. We chose to prime with age in order to direct responses toward the sort of specific, individual attributes that the revised policy might take into account. We chose age specifically because it is a common response to informal iterations of this survey, and in fact is already included in current kidney allocation policy to a certain extent, so we hoped that the skewing effects of the priming would be minimal.

\begin{table}[h!]
    \begin{center}
        \begin{tabular}{ lll } \\\toprule
         Category & Ought & Ought NOT \\ \midrule\midrule
         Age & 80 & 10 \\ \midrule
         Health - Behavioral & 53 & 5 \\ \midrule
         Health - General & 44 & 9 \\ \midrule
         Dependents & 18 & 5 \\ \midrule
         Criminal Record & 9 & 4 \\ \midrule
         Expected Future & 8 & 1 \\ \midrule
         Societal Contribution & 7 & 3 \\ \midrule
         Attitude & 6 & 0 \\ \bottomrule
        \end{tabular}
    \vspace{10pt}
     \caption{\label{tab:Attributes}Categorized responses to the Attribute Collection Survey. The ``Ought'' column counts the number of responses in each category that participants thought should be used to prioritize patients. The ``Ought NOT'' column counts those that participants thought should not be used to prioritize patients. Categories are listed in order of popularity.}
    \end{center}
\end{table}

The three attribute categories that the most participants thought should be used to prioritize patients were ``Age'', ``Health -– Behavioral'' (aspects of health that are generally perceived to be controllable, such as diet and drug use), and ``Health -– General'' (aspects of health that are generally perceived to be involuntary and are unrelated to kidney disease, such as cancer prognosis). There was a sharp drop-off in popularity between the third most popular category, ``Health -- General'' (reported 44 times) and the fourth most popular one, ``Dependents'' (whether the patient had dependents, reported 18 times), so only the first three attribute categories were selected for inclusion in the next stage of the study. The least-commonly reported categories were ``Criminal Record'',  ``Expected Future'', which included responses about patients' future life expectancy and expected quality of life post-surgery, ``Societal Contribution'', and ``Attitude'',  which included responses about patients' psychological state and mental preparation for the surgery and recovery process.

\subsubsection{Participant Demographics}

The survey participants were very diverse, ranging in age from 22 to 64 (with an average age of 40), ranging in self-reported political views from ``extremely liberal'' to ``extremely conservative'', and ranging in educational achievement from ``some high school, no diploma'' to ``doctorate degree''. Participants took between 2 and 26 minutes to complete this survey, with an average completion time of 9 minutes.

\subsection{Evaluating Pairwise Comparisons}

We next gathered data on how people use the three top participant-generated attributes to prioritize patients. We administered a ``Kidney Allocation Survey'' to a new cohort of participants recruited through MTurk.
In this survey, we turned each of the three chosen attributes into a binary one, as described in Table~\ref{tab:Characteristics} below. The Age alternatives represent an adult nearer to the beginning of their adult life (but still of legal drinking age, 30 years old) or nearer to the end (70 years old). \shortened{Of course, for a real implementation we would have to deal with many other ages as well, but we do not address this in this proof-of-concept study.}
For a health-behavioral attribute, we chose alcohol consumption as a (potentially) controllable behavior that can contribute to kidney disease. The indicated amount of alcohol consumption is specified to occur ``prior to diagnosis,'' because drinking afterward disqualifies patients from the waiting list. Skin cancer was chosen as the ``unhealthy'' alternative for the Health-General characteristic because it is a specific, well-known disease that may or may not be fatal and is unrelated to kidney disease.

\begin{table}[h!]
    \begin{center}
        {\small
        \begin{tabular}{ lll } \toprule
         Attribute & Alternative 0 & Alternative 1 \\ \midrule\midrule
         Age  & 30 years old (\textbf{Y}oung) & 70 years old (\textbf{O}ld) \\ \midrule
         Health -\newline Behavioral & 1 alcoholic drink per month (\textbf{R}are) & 5 alcoholic drinks per day (\textbf{F}requent) \\ \midrule
         Health -\newline General & no other major health problems (\textbf{H}ealthy) & skin cancer in remission  (\textbf{C}ancer) \\ \bottomrule
        \end{tabular}
        }
    \vspace{10pt}
    \caption{\label{tab:Characteristics}The two alternatives selected for each attribute. The alternative in each pair that we expected to be preferable was labeled ``0'', and the other was labeled ``1''.}
    \end{center}
\end{table}

Because there are three binary attributes, there are eight possible patient profiles. These eight unique patient profiles were enumerated and assigned ID numbers. For expositional ease, in this paper, we refer to profiles in text as a combination of \{Y,O\}, \{R,F\}, and \{C,H\}, representing \{Young, Old\}, \{Rare, Frequent\} alcohol consumption, and \{Cancer, Healthy\} status. For example, profile YRH reads:

\begin{quote}
    Patient W.A.~is 30 years old, had 1 alcoholic drink per month (prior to diagnosis), and has no other major health problems.
\end{quote}

In the survey, participants were asked to choose between pairs of these profiles. Participants (N = 289) were again recruited through MTurk. They read a short description of how kidney waiting lists work, and were asked to imagine that they were responsible for allocating a single kidney to one of two fictional patients. Each participant was then presented with all ${8 \choose 2} = 28$ possible pairs of profiles, in random order, and asked in each case to select the patient that they believed should receive the kidney. For half of the participants, the profile with the smaller ID number appeared on the screen above the profile with the larger ID number for each question (``original order''), and for the other half of the participants this order was reversed (``reversed order''), to counteract possible ordering or screen location effects.
Each participant received \$1.00 compensation for participating in this part of the study.

\subsubsection{Summary of Responses}

Aggregate responses to the Kidney Allocation Survey are summarized below. The ``Preferred'' column reports 
the percentage of times that each profile
was chosen in all the comparisons in which it appeared.

\begin{table}[h!]
    \begin{center}
        \begin{tabular}{ lllll} \toprule
            Profile & Age & Drinking & Cancer & Preferred \\ \midrule\midrule
            1 (YRH) & 30 & rare & healthy & 94.0\% \\ \midrule
            3 (YRC) & 30 & rare & cancer & 76.8\% \\ \midrule
            2 (YFH) & 30 & frequently & healthy & 63.2\% \\ \midrule
            5 (ORH) & 70 & rare & healthy & 56.1\% \\ \midrule
            4 (YFC) & 30 & frequently & cancer & 43.5\% \\ \midrule
            7 (ORC) & 70 & rare & cancer & 36.3\% \\ \midrule
            6 (OFH) & 70 & frequently & healthy & 23.6\% \\ \midrule
            8 (OFC) & 70 & frequently & cancer & 6.4\% \\ \bottomrule
        \end{tabular}
        \vspace{10pt}
        \caption{\label{tab:Ranking}Profile ranking according to Kidney Allocation Survey responses. The ``Preferred'' column describes the percentage of time the indicated profile was chosen among all the times it appeared in a comparison.}
    \end{center}
\end{table}

As expected, there was a clear preference for profile 1 (30 years old, 1 alcoholic drink per month, no other major health problems), and a clear preference against profile 8 (70 years old, 5 alcoholic drinks per day, skin cancer in remission). 

The preference for profile 3 (skin cancer in remission but minimal drinking) over profile 2 (healthy other than heavy drinking), and similarly 7 over 6, suggests that participants put greater weight on the health-behavioral attribute than on the health-general one. This aligns with responses to our first survey, in which more participants gave responses in the ``Health - Behavioral'' category than gave responses in the ``Health - General'' category (see Table~\ref{tab:Attributes}). (Of course, this observation may not generalize to other health-behavioral and health-general attributes, such as drinking soda and skin cancer that's not in remission.)

\subsubsection{Participant Demographics}

Again, the survey participants were very diverse. They ranged in age from 19 to 70 (with an average age of 37), and again ranged in self-reported political views from ``extremely liberal'' to ``extremely conservative'', and in educational achievement from ``some high school, no diploma'' to ``doctorate degree''. Participants took between 1.5 minutes and 38.5 minutes to complete this survey, with an average completion time of 7 minutes.

\subsection{Estimating Profile Scores}

We performed statistical modeling of participants' pairwise comparisons between patient profiles
in order to obtain weights for each profile. We used the Bradley-Terry model, which treats each pairwise comparison as a contest between a pair of players~\cite{BRADLEY1984299}.
Under this model, each player $i$ has a score $p_i$, representing its skill or value. Given two players $i$ and $j$ with respective scores $p_i$ and $p_j$, the probability that player $i$ will win the contest is:

\[ P(i>j) = \frac{p_i}{p_i+p_j} \]

To illustrate this model, imagine that individuals $a$, $b$, and $c$ are patients waiting for kidney transplants. For each pair of patients, imagine that we have asked 100 survey participants to pick one to receive a kidney. Assume that patient $a$ was picked over patient $b$ 63 times and picked over patient $c$ 72 times, and that patient $b$ was picked over patient $c$ 58 times. There were no ties. We can use the Bradley-Terry model to estimate a score representing the value that survey participants place on giving a kidney to each patient. 
Judging from this sample of 300 comparisons, the probability of patient $a$ being chosen over patient $b$ is 63/100 = 0.63, the probability of patient $a$ being chosen over patient $c$ is 72/100 = 0.72, and the probability of patient $b$ being chosen over patient $c$ is 58/100 = 0.58. Therefore, we have:
\[ P(a>b) = 0.63 = \frac{p_a}{p_a+p_b} \]
\[ P(a>c) = 0.72 = \frac{p_a}{p_a+p_c} \]
\[ P(b>c) = 0.58 = \frac{p_b}{p_b+p_c} \]

When we fit the model to these results and assign score $1.00$ to $p_a$, $p_b$ is estimated as $0.57$, and $p_c$ is estimated as $0.40$. It is important to note that these scores are only meaningful relative to each other.  In particular, scaling all the scores $p_i$ by the same factor would not affect the predictions.

Based on these scores, the preference probabilities are estimated as follows:
\[ P(a>b) = \frac{p_a}{p_a+p_b} = \frac{1.00}{1.00+0.57} \approx 0.64  \]
\[ P(a>c) = \frac{p_a}{p_a+p_c} = \frac{1.00}{1.00+0.40} \approx 0.71  \]
\[ P(b>c) = \frac{p_b}{p_b+p_c} = \frac{0.57}{0.57+0.40} \approx 0.59  \]
Note that these scores do not exactly line up with the empirical fractions with which each patient is chosen ($0.63$, $0.72$, and $0.58$, respectively); this is because we only have 2 degrees of freedom.   Specifically, if we decrease $p_a$ then the estimate gets closer to the first empirical fraction but further away from the second; if we decrease $p_b$ the estimate gets closer to the third but further away from the first; and if we decrease $p_c$ the estimate gets closer to the second but further away from the third.

The BT scores (that we estimate based on our data) constitute one measure of the value that the survey participants collectively place on ``saving'' each profile. The higher this value, the more likely a randomly selected participant is to select that profile over another.  We can then use these scores as weights.  (One may wonder whether perhaps it would be better to somehow transform---e.g., take the  square root of---the weights first; one of our experiments below suggests this would make almost no difference.)
This estimation procedure constitutes a specific way to {\em aggregate} the human subjects' moral judgments into a single weight for each profile; the strategy of using social choice theory to aggregate moral preferences for decision making has already been proposed by several groups~\cite{Greene16:Embedding,Conitzer17:Moral,Noothigattu18:Voting}, and our specific approach fits well in the literature on interpreting voting as a method for statistically estimating an underlying truth (for an overview, see~\citet{ElkindSlinkoChapter}).

We estimate BT scores in two different ways.  One is to estimate scores directly for all profiles, so one profile's score is not constrained by the scores of other profiles.  The second is to consider the importance of the individual attributes and let the score of profile $i$ be a linear function of these:
\[ \sum_{r=1}^{p} \beta_{r}x_{ir}+U_{i}\]
where $x_{ir}$ is profile $i$'s value for attribute $r$, and we estimate the $\beta_r$ (importance of attribute $r$). The $U_i$ are individual error terms where $U_i \sim N(0,\sigma^2)$, resulting in correlation between comparisons that share a common profile.

We used the \texttt{BTm()} function in the BradleyTerry2 package in R to estimate profile scores $p_1, \ldots, p_8$ based on the 8092 pairwise comparisons, both directly and as a function of the estimated scores of their three attribute values. The most-preferred profile, profile 1 in both cases, was assigned a score of 1.  The results are in Table~\ref{tab:BT-scores}.

\begin{table}[h!]
    \begin{center}
        \begin{tabular}{ lll } \toprule
             Profile & Direct & Attribute-based \\ \midrule\midrule
             1 (YRH) & 1.000000000 & 1.00000000 \\ \midrule
             3 (YRC) & 0.236280167 & 0.13183083 \\ \midrule
             2 (YFH) & 0.103243396 & 0.29106507 \\ \midrule
             5 (ORH) & 0.070045054 & 0.03837135 \\ \midrule
             4 (YFC) & 0.035722844 & 0.08900390 \\ \midrule
             7 (ORC) & 0.024072427 & 0.01173346 \\ \midrule
             6 (OFH) & 0.011349772 & 0.02590593 \\ \midrule
             8 (OFC) & 0.002769801 & 0.00341520 \\ \bottomrule
        \end{tabular}
    \vspace{10pt}
    \caption{\label{tab:BT-scores}The patient profile scores estimated using the Bradley-Terry Model. The ``Direct'' scores correspond to allowing a separate parameter for each profile (we use these in our simulations below), and the ``Attribute-based'' scores are based on the attributes via the linear model.}
    \end{center}
\end{table}

\subsection{Adapting the Algorithm}

The final step was to incorporate the obtained weights into the kidney exchange market clearing algorithm. 
Because our human subject data and analysis do not involve comparisons between differing quantities of patient profiles (e.g., choosing two patients with profile 1 over three patients with profile 2), we feel it is inappropriate to use the weights for such decisions.
We only use the weights to break ties between solutions of maximum cardinality.

To find a matching, our adapted (prioritized) algorithm first runs the basic IP-based algorithm due to~\citet{Abraham07:Clearing} with unit edge weights (i.e., $w_e = 1 \ \forall e \in E$). Given a pool of patient-donor pairs, this algorithm returns a set of kidney exchange cycles that maximizes the number of patients who receive a kidney without regard to their personal characteristics (other than medical compatibility). Our algorithm records the number of patients who receive a kidney in this solution as $Q$, and adds a new constraint to the IP requiring that the solution includes at least $Q$ vertices. We then re-solve the IP with a new objective, using the weights corresponding to the patient profile scores derived from the survey responses.  Formally, with $|c|$ denoting the \emph{number} of vertices in cycle $c$,  $\ptype : V \to \{1,\ldots,8\}$ mapping a vertex to its patient's profile, and $w_\theta$ denoting the score of profile $\theta$, we solve:

$$
\begin{array}{lll}
\max & \sum_{c \in C(L,K)} \left[ \sum_{(u,v) \in c} w_{\ptype(v)} \right] \ x_c & \\
\text{s.t.} & \sum_{c : v \in c} x_c \leq 1 & \forall v \in V \\
& \sum_{c \in C(L,K)} |c| x_c \geq Q & \\
& x_c \in \{0,1\} & \forall c \in C(L,K) \\
\end{array}
$$
This results in a set of kidney exchange cycles that includes the maximum possible number of patients, but prioritizes patient profiles that the surveyed population preferred.

\section{Experiments}

Having described how we obtained weights and how we integrated these weights into the IP-based algorithm,
we now describe our experiments testing the effects of our prioritizing algorithm in simulations.

\subsection{Experimental Setup}

Based on previously developed tools~\cite{Dickerson15:FutureMatch},
we built a simulator to mimic daily matching in a real\cutout{-world} kidney exchange pool.\footnote{All code for this paper can be found in the \texttt{Ethics} package of \texttt{github.com/RachelFreedman/KidneyExchange}.} In the simulation, each day, some incompatible patient-donor pairs enter the simulated pool and some depart.  
Then, a matching algorithm is run to match a subset of compatible patient-donor pairs. 
\cutout{The remaining}\shortened{Remaining} incompatible pairs stay in the pool for consideration on the next day (and possibly beyond). Finally,\cutout{ the} matches formed the previous day are executed with a certain success probability, and\cutout{ the} matched pairs are removed from the pool. Not all of the matched pairs are executed, because in real-life situations many algorithmic matches fail to go to transplant due to last-minute medical incompatibilities, surgeons rejecting a donor organ, or other logistical difficulties~\cite{Anderson15:Finding,Dickerson18:Failure} We model this by executing matches with a probability of 0.5 instead of 1.

The demographics of our simulated pool were designed to reflect the UNOS kidney exchange pool where possible, and otherwise the general US population.

\subsection{Experiment 1: Matchings with pair scores}

\subsubsection{Experiment} In the first experiment, we compared the patient-donor pairs (vertices) matched by the original algorithm, which treats all profiles equally and breaks ties arbitrarily, to the pairs matched by the ``prioritized'' algorithm, which breaks ties towards pairs with higher (patient) profile scores. We ran 20 simulations of daily matching over the course of 5 simulated years using both algorithms.

\subsubsection{Hypothesis}

We hypothesized that the original algorithm would match pairs in approximately the same proportion for every profile, but that the prioritizing algorithm would match pairs with higher profile scores more often than pairs with lower scores. Moreover, we hypothesized that the pairs with the highest profile scores (profiles 1, 3, and 2) would be matched more often by the prioritizing algorithm than by the original algorithm, and that the pairs with the lowest profile scores (profiles 7, 6, and 8) would be matched more often by the original algorithm than the prioritizing algorithm.

\subsubsection{Results}

The proportions of pairs of each profile type matched by the original and prioritizing algorithms are plotted in Figure~\ref{fi:prop} below. ``Proportion Matched'' is the proportion of pairs that entered the pool that were subsequently matched. Both algorithms matched approximately 61.7\% of pairs overall.  (This result does not follow immediately from the fact that both algorithms match the maximum number of pairs in each round, because which specific profiles are matched in a round will affect which profiles appear in future rounds, and consequently may affect how many can be matched in future rounds.)

The results support both of our hypotheses. First, the original algorithm, called ``STANDARD'' in Figure~\ref{fi:prop}, matched pairs approximately 62\% of the time, regardless of their profile, while the prioritizing algorithm, called ``PRIORITIZED''  in Figure~\ref{fi:prop}, matched the pairs with profile 1, who had the highest profile scores, nearly twice as often as it matched pairs with profile 8, who had the lowest profile scores.

Secondly, pairs with profiles 1, 3, and 2 were indeed matched substantially more often by the prioritizing algorithm than by the original algorithm, while pairs with profiles 7, 6, and 8 were indeed matched substantially less often by the prioritizing algorithm than by the original algorithm. Thus, the scores assigned by the prioritizing algorithm do have a substantial effect on which profiles get matched.

\begin{figure}[h!]
  \centering
  \includegraphics[width=0.9\textwidth]{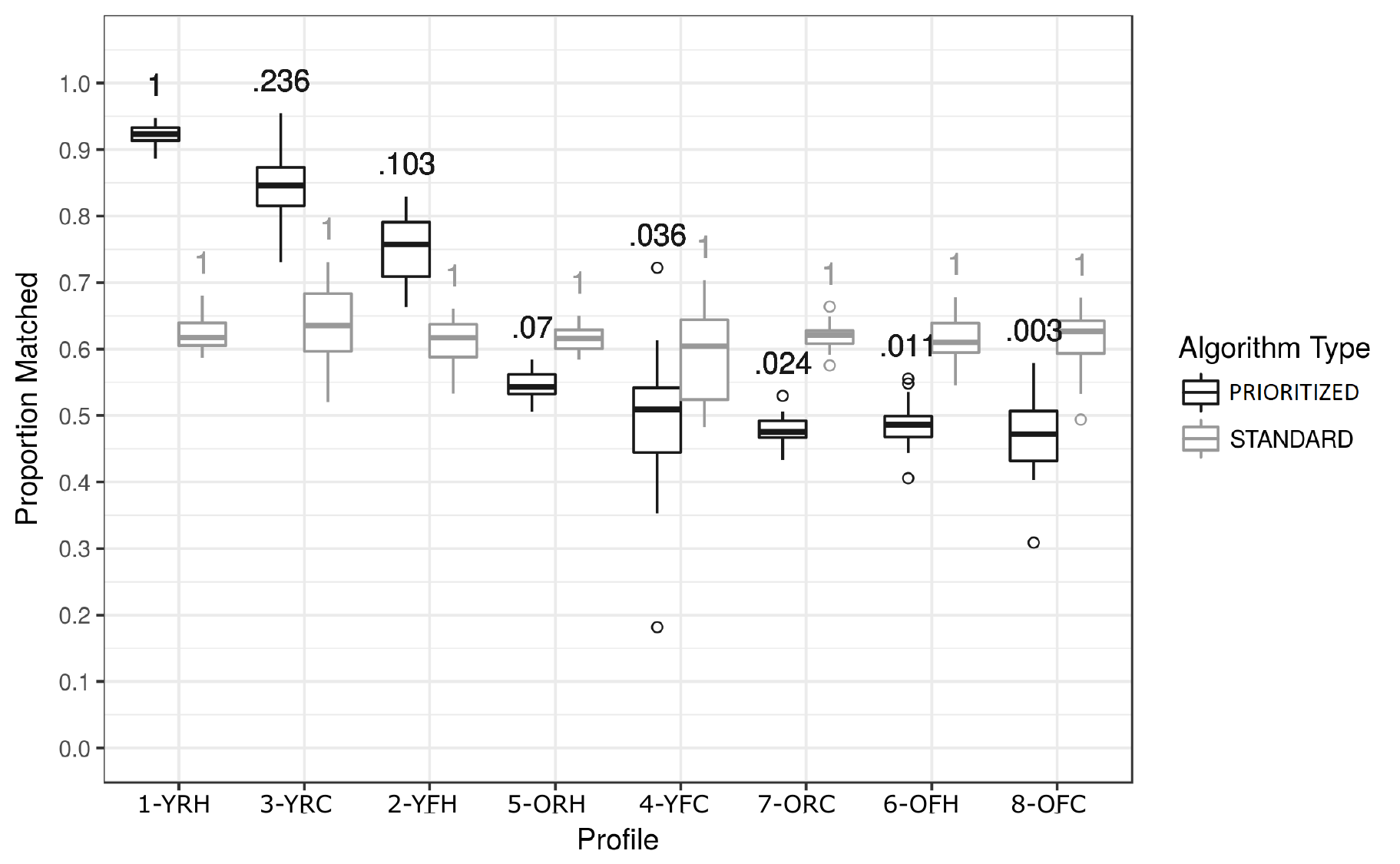} 
   \caption{The proportions of pairs matched over the course of the simulation, by profile type and algorithm type. N = 20 runs were used for each box. The numbers are the scores assigned (for tiebreaking) to each profile by each algorithm type. Because the STANDARD algorithm treats all profiles equally, it assigns each profile a score of 1. In this figure and later figures, each box represents the interquartile range (middle 50\%), with the inner line denoting the median. The whiskers extend to the furthest data points within 1.5 $\times$ the interquartile range of the median, and the small circles denote outliers beyond this range.}
\label{fi:prop}
\end{figure}

\subsection{Experiment 2: Matchings evaluated by blood type}

\subsubsection{Experiment} Blood type is a major factor in determining patient-donor biological compatibility (see Table~\ref{tab:Compatibility} for a summary of blood type compatibility). Patients with difficult-to-match blood types are more likely to struggle to find a compatible donor, and consequently can be disproportionately represented in kidney exchange pools. To explore how the modified algorithm treats patients with these blood types, we again ran 20 simulations of 5 simulated years of daily matching, this time recording the patient and donor blood types of each pair in addition to their profiles. We partitioned pairs into four established blood type classes motivated by large market analysis~\cite{Ashlagi14:Free,Toulis15:Design}. {\em Underdemanded} pairs were those that contain a patient with blood type O, a donor with blood type AB, or both, making them the most difficult to match. {\em Overdemanded} pairs contain a patient with blood type AB, a donor with blood type O, or both; {\em self-demanded} pairs contain a patient and donor with the same blood type; and {\em reciprocally demanded} pairs contain one person with blood type A, and one person with blood type B. These three classes are substantially easier to match. 

\subsubsection{Hypothesis}

We hypothesized that the prioritizing algorithm primarily impacts underdemanded pairs, prioritizing underdemanded pairs with higher profile scores at the expense of underdemanded pairs with lower profile scores, while matching pairs that belong to the three other blood type classes at roughly the same high rates that the original algorithm does.  The reasoning was that, intuitively, there is generally a scarcity of matching opportunities for the underdemanded pairs, but this is not so for the other types of pairs.

\subsubsection{Results} 
\begin{figure}[h!]
  \centering
  \includegraphics[width=0.9\textwidth]{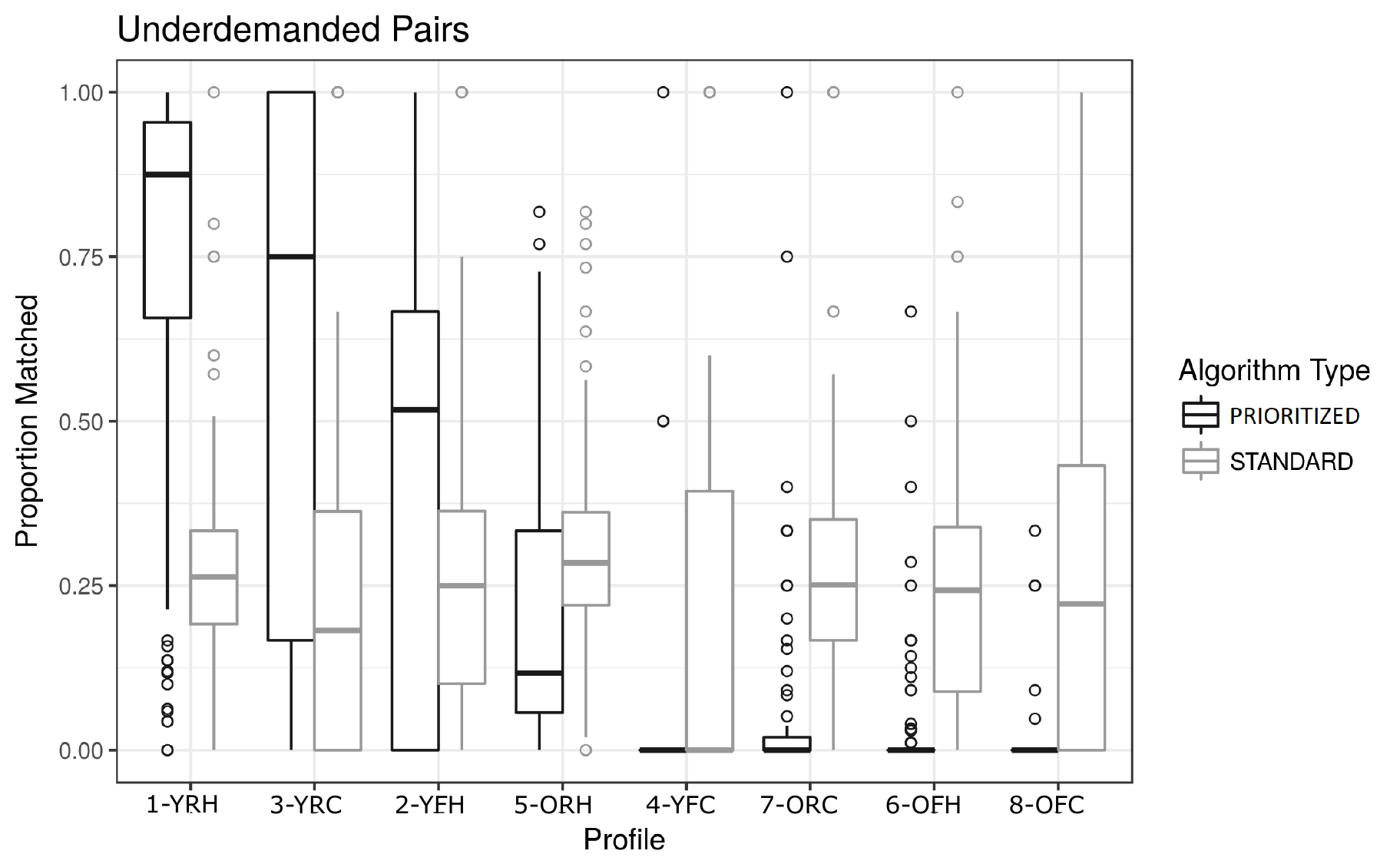}
  \caption{The proportions of underdemanded pairs matched over the course of the simulation, by profile type and algorithm type. N = 20 runs were used for each box.}
\label{fi:under}
\end{figure}

\begin{figure}[h!]
  \centering
  \includegraphics[width=0.9\textwidth]{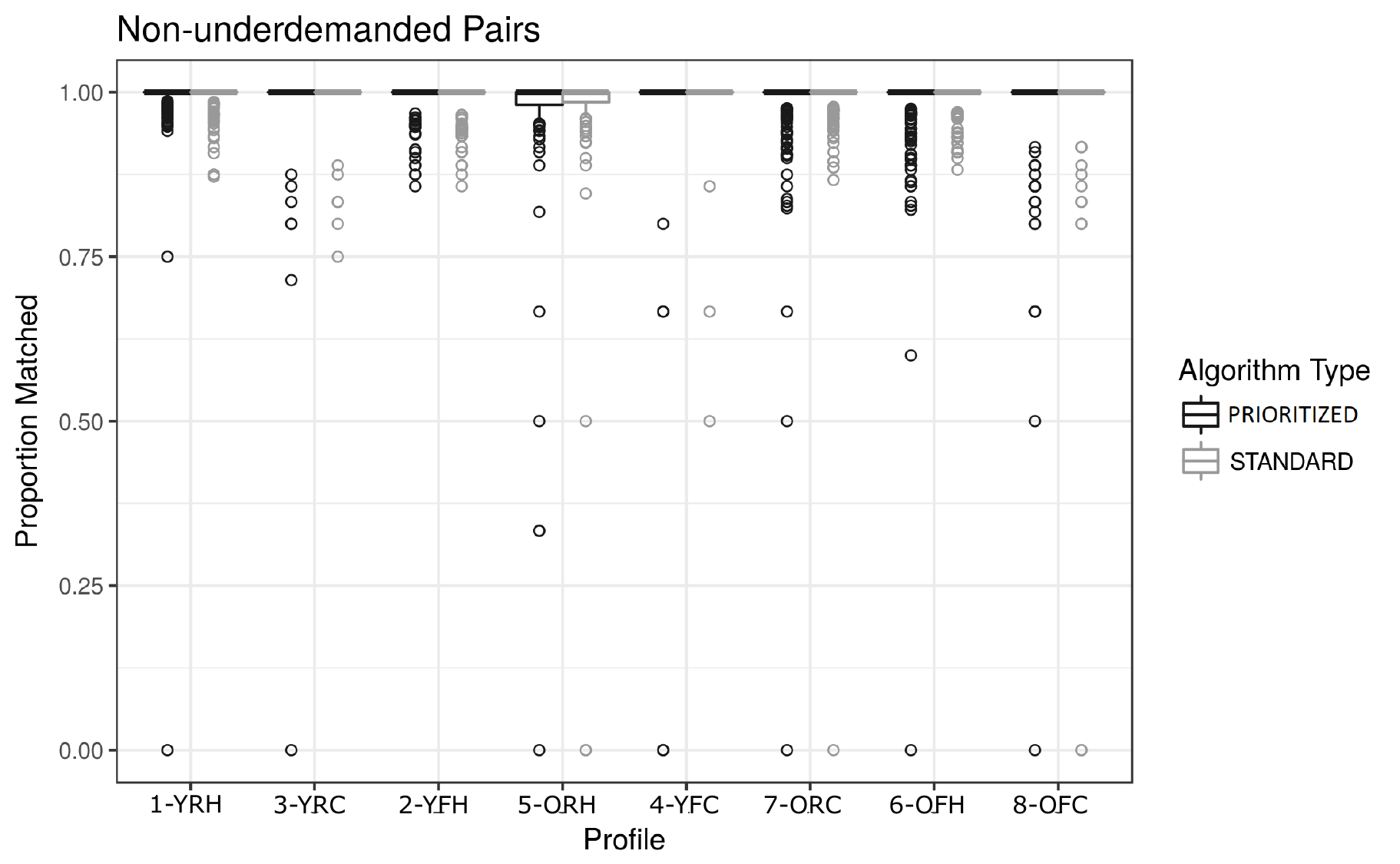} 
  \caption{The proportions of overdemanded, self-demanded, and reciprocally demanded pairs grouped together matched over the course of the simulation, by profile type and algorithm type. N = 60 runs were used for each box.}
  \label{fi:notunder}
\end{figure}

The results confirm our hypothesis.  The proportions of underdemanded pairs matched are plotted in Figure~\ref{fi:under}. We found the proportions of overdemanded, self-demanded, and reciprocally demanded profiles matched to be fairly similar, so we grouped them together in Figure~\ref{fi:notunder}.  The prioritizing algorithm matched underdemanded pairs with high profile scores substantially more often and underdemanded pairs with low scores substantially less often than the original algorithm did, but both algorithms matched pairs of other classes at roughly equal rates. This suggests that the primary difference between the algorithms lies in how they treat underdemanded pairs.

\subsection{Experiment 3: Transforming Bradley-Terry scores }

\subsubsection{Experiment} 
One may well wonder whether using the Bradley-Terry scores as weights is well motivated, especially because the difference in scores between the top two profiles is so large. This difference reflects that it is very unlikely that the top profile would not be preferred by a subject, but this does not imply that saving someone of profile 1 is more than four times as important as saving someone of profile 3.
Presumably, the ideal weights used in the algorithm would be monotonically increasing in the BT scores, but it is not clear that they should be proportional.  To explore the impact of the sizes of the gaps between the weights on the matchings produced by the PRIORITIZED algorithm, we tried alternative weights, given below.

\begin{table}[h]
    \begin{center}
    \label{scores}
    {\small
    \begin{tabular}{lllllllll} \toprule
        & & & & Profile & & & & \\
         & 1 & 2 & 3 & 4 & 5 & 6 & 7 & 8 \\ \midrule\midrule
        ORIGINAL & 1 & .103 & .236 & .036 & .070 & .011 & .024 & .003 \\ \midrule
        LINEAR & 1 & .998 & .999 & .996 & .997 & .994 & .995 & .993  \\ \bottomrule
    \end{tabular}}
    \vspace{10pt}
    \caption{Two weight vectors.  The first represents the original BT scores as used in PRIORITIZED; the second agrees with the BT scores on the ordering, but the weights are linear in the rank of the profile, as used in LINEAR PRIORITIZED.}
    \end{center}
\end{table}

\noindent The alternative weights result in the profiles being ranked in the same order as the BT scores, but make the difference between sequential weights small and identical. 

We again ran 20 simulations of 5 simulated years of daily matching, this time comparing the prioritized algorithm using the original BT scores as weights to the prioritized algorithm using the alternative weights. 

\subsubsection{Hypothesis}

We hypothesized that the profile {\em ranking} was primarily responsible for the differences in matching and that beyond this, the magnitude of the BT scores would not have a great impact.

Hence, since both of these vectors of weights rank profiles the same, we expected them to match profiles in very similar proportions.

\subsubsection{Results} 
\begin{figure}[h!]
  \centering
  \includegraphics[width=0.9\textwidth]{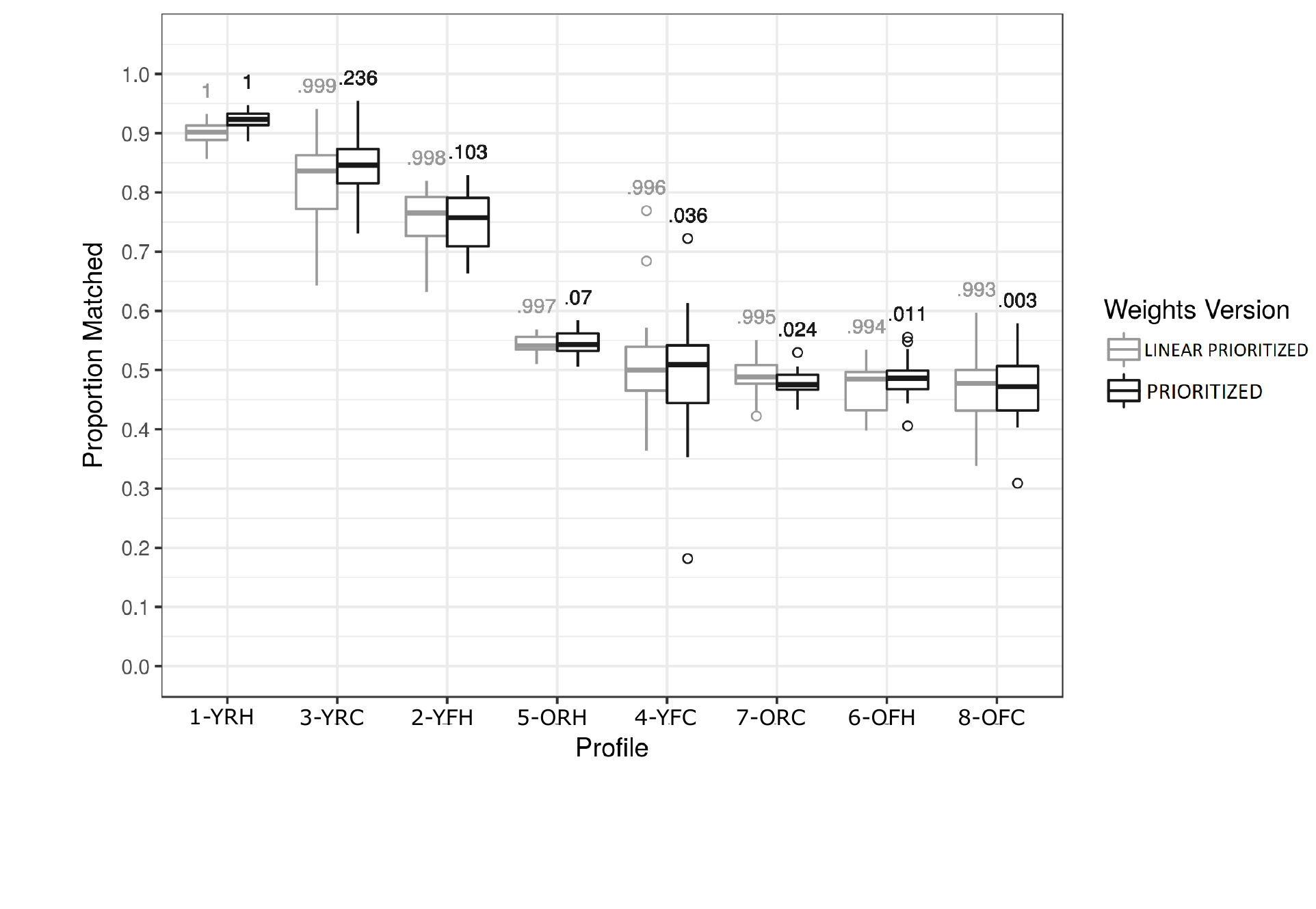}
  \vspace{-20pt}
  \caption{The proportions of underdemanded pairs matched over the course of the simulation, by profile and algorithm. The ``PRIORITIZED'' algorithm matches using the original profile weights, while the ``LINEAR PRIORITIZED'' algorithm matches using the alternative weights  given above.}
  \label{fi:linear}
\end{figure}

The proportions of pairs matched using each weight vector are plotted in Figure~\ref{fi:linear}. The matching using the original weights is again called ``PRIORITIZED'', while the matching using the new weight vector is called ``LINEAR PRIORITIZED''. The results confirm our hypothesis. There was very little difference in the matchings produced by the PRIORITIZED and LINEAR PRIORITIZED algorithms, and what difference there was could be easily explained by the fact that a slightly different set of pairs enter the pool for each algorithm type.
We also tried other weight vectors that assigned different weights to each profile, but that agreed with the initial prioritizing algorithm on the order of the profiles, and found similarly little difference.
These results suggest that the profile ranking induced by the weights is primarily responsible for the impact of the prioritizing algorithm, while beyond that varying the weights makes little difference.

\cutout{

\section{Discussion}

In this section, we discuss the potential for applying these results to real-world kidney exchanges, some of the ethical context of our work, and directions for future research.

\subsection{Application in Real Kidney Exchanges}

Our study serves as a proof-of-concept for the proposed method of soliciting and using prioritization weights, but we do not advocate directly applying the weights obtained in our limited study to a real kidney exchange.  For one, a real kidney exchange would require each of the attributes
considered to be able to take more possible values than we tested in our mere pairwise comparisons (e.g., there should be more than two values for ``age'').    Whoever eventually makes the judgments about who should be prioritized (in our study this was left to MTurkers, who may not be representative of the general population)
should also have a chance to obtain expert advice---for example, about what the prognosis is for someone with skin cancer in remission.  Generally, deploying these techniques in a real kidney exchange should be done with input from representatives of all the stakeholders
in such a system---patients, donors, surgeons, other hospital staff, etc. How to best structure the process as a whole is an important topic for future research.

That being said, our work demonstrates that there are no fundamental technical obstacles to building such a system.  We have shown one way in which
moral judgments can be elicited from human subjects, how those judgments can be statistically modeled, and how the results can be incorporated into the algorithm.  We have also shown, through simulations, what the likely effects of deploying such a prioritization system would be, namely
that underdemanded pairs would be significantly impacted but little would change for others.  We 
do not make any judgment about whether this conclusion speaks in favor of or against such prioritization, but expect the conclusion to be robust to changes in the prioritization such as those that would result from a more thorough process, as described in the previous paragraph.  We also expect the conclusion to hold if the method is applied to real rather than simulated data: while the distribution of donor and patient data in real kidney exchanges
is surely different from the simulated one, there are no obvious reasons to suspect that this would change our qualitative conclusion.

\subsection{Artificial Morality}

Our work is also a concrete proof-of-concept of a hybrid approach to artificial morality. This hybrid combines and contrasts with both top-down approaches and bottom-up approaches~\cite{Allen05:Artificial, Wallach10:Moral}.

Top-down approaches provide a computer with a general ethical theory along with facts that are morally relevant according to the theory. The machine then infers moral judgments or makes moral decisions by applying the theory to the facts. This top-down approach must begin by choosing a moral theory to program into computers. The problem is that ethicists support a wide variety of moral theories, and it is hard to see how to justify insisting on one theory instead of another. The second problem for the top-down approach is that such theories are too vague to implement, they conflict in real-life decisions, and they can yield disasters~\cite{Allen05:Artificial, Wallach10:Moral, Anderson11:Machine, Pereira17:Agent, Bringsjord18:Contextual}. It might seem innocuous for a computer to follow a rule like ``Minimize harm,'' but what if a computer decides that killing all humans will minimize harm in the long run? 

Bottom-up approaches try to avoid assuming any moral theory by using machine learning trained on human descriptions of concrete moral problems to predict human moral judgments. This system mirrors one way in which children learn morality, so it resembles Alan Turing's original proposal for developing artificial intelligence. Over 50 years ago, when faced with the problem of developing an artificial agent capable of making decisions like an adult human, Turing presciently suggested, ``Instead of trying to produce a programme to simulate the adult mind, why not rather try to produce one which simulates the child's? If this were then subjected to an appropriate course of education, one would obtain the adult brain.''~\cite{Turing50:Computing} It is not completely clear how children learn morality, but one element involves encounters with concrete moral problems. Accordingly, a purely bottom-up approach might try to build AI systems that can learn solely by exposing them to moral examples.

In order to learn in this way, an artificial moral judge or agent must be able to safely extrapolate what it learned in some cases to novel environments that it may face in the future. This ambitious purpose requires a data set that is large and varied enough to teach genuine underlying moral principles that are projectible into these new environments as opposed to narrowly applicable surface features that predict human moral judgments only in the training set. Moreover, we want to know not only that an act is wrong but also why it is wrong---what makes it wrong. Otherwise, we cannot give comprehensible justifications for controversial decisions, and we have no way to check on whether or when the system is working properly. These goals are difficult or impossible to achieve within uninterpretable ML systems~\cite{Baum18:From}.  Moreover, today's AI systems lack a broad / commonsense understanding of our (human) world, and it seems that such an understanding would be a necessary component of a system that could make moral judgments across a broad range of settings. This task could be aided by computational models of case-based and value-based reasoning and argumentation. For a variety of approaches, see \citet{Rahwan09:Argumentation}.

While top-down and bottom-up approaches both contain promising elements for developing moral artificial intelligence, each approach also faces serious challenges~\cite{Baum18:From}. Their flaws suggest that we might be able to do better by combining the two approaches into unified systems that achieve the benefits without the problems of each~\cite{Conitzer17:Moral}. Top-down supervision and organization can enable artificial agents to justify their decisions in terms of moral principles that are comprehensible to humans, while bottom-up learning has the potential to deal with complex facts in particular cases. 

Our particular hybrid attempts to reduce the arbitrariness of top-down approaches by crowd-sourcing a list of features that humans see as morally relevant. Humans do not include some characteristics, such as shirt color, as relevant to kidney exchanges, and they are responsible for determining which patient characteristics are important. These features of alternatives can then be used both to constrain the data and also to provide an interpretable basis for the algorithm's predictions. In this way, our hybrid introduces some minimal theory in the form of morally relevant features in order to solve the main problems of competing top-down and bottom-up approaches. 

This hybrid method is particularly well-suited to developing ethical machine reasoning in constrained domains where it is clear which features of acts are morally relevant. 
In such a domain, it is possible to create models of multiple individuals' moral decisions, and then to have these models {\em vote} over what the right decision is overall~\citep{Conitzer17:Moral}.
\citet{Noothigattu18:Voting} recently applied a version of this approach to ethical decision-making for autonomous vehicles. They aggregated human moral judgments about autonomous vehicles colliding with, for example, a pedestrian (likely killing them) or a wall (likely killing the driver). Their method assumes that causing death along with a few other features are morally relevant. 

We applied a similar approach to kidney allocation in this paper. Our hybrid approach is not without its own challenges, however. One is that human moral judgments are inconsistent within and across individuals, so a machine learning system can at best predict a subset (though perhaps a majority) of human moral judgments. We will need to decide how to make social decisions in light of such disagreements. 

Moreover, humans often exhibit biases, such as racial and gender discrimination, that they themselves reject as improper and would want an artificial moral agent to avoid. This problem can be reduced (though not fully solved) by designing the artificial intelligence system to include only features that most humans deem to be morally relevant. If we had included the characteristic ``race'' in our patient descriptions, the algorithm might have learned to take race into account. Leaving out that characteristic avoids this undesirable result, though it still leaves open the possibility of more subtle and hidden forms of bias. These problems for our hybrid approach will be the topic of future work.
 
\subsection{Future Research}

Besides being applicable to kidney (and perhaps other organ) exchanges, our study also suggests a roadmap for automated moral decision making in other domains.  For example, the idea of obtaining human subjects' judgments to guide AI systems in moral decision making is also being explored for self-driving cars~\cite{Bonnefon16:Social,Noothigattu18:Voting}.  Some aspects of that domain are different.  In particular, in that case the need for automated decision-making is driven by the fact that decisions need to be made too fast to be made by a human, whereas in kidney exchanges the need for AI is driven by the fact that the nature of the search space of all possible matchings makes the problem intractable for a human.  Nevertheless, the domains clearly have much in common, and it seems likely that we will be confronted with similar problems in many others.
Further research should eventually lead us to a good understanding of best practices for automated moral decision making by generalizing from human judgments.
}

\subsubsection{Acknowledgments}
This  work  is  partially  supported  by  the  project  ``How to Build Ethics into Robust Artificial Intelligence'' funded by the  Future  of  Life  Institute, by the Templeton World Charity Foundation, by NSF IIS-1527434, and by Duke Bass Connections. Dickerson was supported in part by NSF CAREER Award IIS-1846237 and a Google Faculty Research Award. Conitzer was supported in part by NSF IIS-1814056. We thank Lirong Xia, Zhibing Zhao, and Kyle Burris, and members of our moral AI group at Duke, including Yuan Deng, Kenzie Doyle, Jeremy Fox, Max Kramer, and Eitan Sapiro-Gheiler, for feedback on this work.  

\newpage

\bibliographystyle{plainnat}  
\bibliography{references} 

\begin{thebibliography}{46}
\providecommand{\natexlab}[1]{#1}
\providecommand{\url}[1]{\texttt{#1}}
\expandafter\ifx\csname urlstyle\endcsname\relax
  \providecommand{\doi}[1]{doi: #1}\else
  \providecommand{\doi}{doi: \begingroup \urlstyle{rm}\Url}\fi

\bibitem[Abraham et~al.(2007)Abraham, Blum, and Sandholm]{Abraham07:Clearing}
David Abraham, Avrim Blum, and Tuomas Sandholm.
\newblock Clearing algorithms for barter exchange markets: {E}nabling
  nationwide kidney exchanges.
\newblock In \emph{ACM EC}, pages 295--304, 2007.

\bibitem[Allen et~al.(2005)Allen, Smit, and Wallach]{Allen05:Artificial}
Colin Allen, Iva Smit, and Wendell Wallach.
\newblock Artificial morality: Top-down, bottom-up, and hybrid approaches.
\newblock \emph{Ethics and Inf. Technol.}, 7\penalty0 (3):\penalty0 149--155,
  September 2005.
\newblock ISSN 1388-1957.

\bibitem[Anderson and Anderson(2011)]{Anderson11:Machine}
M.~Anderson and S.~Anderson.
\newblock \emph{Machine Ethics}, pages 231--492.
\newblock Cambridge Univ. Press, 2011.

\bibitem[Anderson et~al.(2015)Anderson, Ashlagi, Gamarnik, and
  Roth]{Anderson15:Finding}
Ross Anderson, Itai Ashlagi, David Gamarnik, and Alvin~E Roth.
\newblock Finding long chains in kidney exchange using the traveling salesman
  problem.
\newblock \emph{PNAS}, 112\penalty0 (3):\penalty0 663--668, 2015.

\bibitem[Ashlagi and Roth(2014)]{Ashlagi14:Free}
Itai Ashlagi and Alvin~E. Roth.
\newblock Free riding and participation in large scale, multi-hospital kidney
  exchange.
\newblock \emph{Theoretical Economics}, 9:\penalty0 817--865, 2014.

\bibitem[Ashlagi et~al.(2017)Ashlagi, Gamarnik, Rees, and Roth]{Ashlagi17:Need}
Itai Ashlagi, David Gamarnik, Michael Rees, and Alvin~E. Roth.
\newblock The need for (long) chains in kidney exchange, 2017.
\newblock Initial version appeared at the ACM Conference on Electronic Commerce
  (EC-12).

\bibitem[Barnhart et~al.(1998)Barnhart, Johnson, Nemhauser, Savelsbergh, and
  Vance]{Barnhart98:Branch-and-Price}
Cynthia Barnhart, Ellis~L. Johnson, George~L. Nemhauser, Martin W.~P.
  Savelsbergh, and Pamela~H. Vance.
\newblock Branch-and-price: Column generation for solving huge integer
  programs.
\newblock \emph{Operations Research}, 46\penalty0 (3):\penalty0 316--329, 1998.

\bibitem[Baum et~al.(2018)Baum, Hermanns, and Speith]{Baum18:From}
Kevin Baum, Holger Hermanns, and Timo Speith.
\newblock From machine ethics to machine explainability and back.
\newblock In \emph{{ISAIM}}, 2018.

\bibitem[Bir\'{o} and Cechl\'{a}rov\'{a}(2007)]{Biro07:Inapproximability}
P{\'e}ter Bir\'{o} and Katar\'{\i}na Cechl\'{a}rov\'{a}.
\newblock Inapproximability of the kidney exchange problem.
\newblock \emph{Information Processing Letters}, 101\penalty0 (5):\penalty0
  199, March 2007.

\bibitem[Bir{\'o} et~al.(2009)Bir{\'o}, Manlove, and Rizzi]{Biro09:Maximum}
P{\'e}ter Bir{\'o}, David~F Manlove, and Romeo Rizzi.
\newblock Maximum weight cycle packing in directed graphs, with application to
  kidney exchange programs.
\newblock \emph{Discrete Mathematics, Algorithms and Applications}, 1\penalty0
  (04):\penalty0 499--517, 2009.

\bibitem[Bir{\'o} et~al.(2017)Bir{\'o}, Burnapp, Haase, Hemke, Johnson, van~de
  Klundert, and Manlove]{Biro17:Kidney}
P{\'e}ter Bir{\'o}, Lisa Burnapp, Bernadette Haase, Aline Hemke, Rachel
  Johnson, Joris van~de Klundert, and David Manlove.
\newblock Kidney exchange practices in {E}urope, 2017.
\newblock First Handbook of the {COST} {A}ction {CA}15210: {E}uropean {N}etwork
  for {C}ollaboration on {K}idney {E}xchange {P}rogrammes.

\bibitem[Blum et~al.(2017)Blum, Caragiannis, Haghtalab, Procaccia, Procaccia,
  and Vaish]{Blum17:Opting}
Avrim Blum, Ioannis Caragiannis, Nika Haghtalab, Ariel Procaccia, Eviatar
  Procaccia, and Rohit Vaish.
\newblock Opting into optimal matchings.
\newblock In \emph{SODA}, 2017.

\bibitem[Bonnefon et~al.(2016)Bonnefon, Shariff, and Rahwan]{Bonnefon16:Social}
Jean-Fran\c{c}ois Bonnefon, Azim Shariff, and Iyad Rahwan.
\newblock The social dilemma of autonomous vehicles.
\newblock \emph{Science}, 352\penalty0 (6293):\penalty0 1573--1576, June 2016.

\bibitem[Bradley(1984)]{BRADLEY1984299}
Ralph~A. Bradley.
\newblock 14 paired comparisons: Some basic procedures and examples.
\newblock \emph{Handbook of Statistics}, 4:\penalty0 299 -- 326, 1984.
\newblock ISSN 0169-7161.
\newblock \doi{http://dx.doi.org/10.1016/S0169-7161(84)04016-5}.
\newblock URL
  \url{http://www.sciencedirect.com/science/article/pii/S0169716184040165}.
\newblock Nonparametric Methods.

\bibitem[Bringsjord et~al.(2018)Bringsjord, G., Malle, and
  Scheutz]{Bringsjord18:Contextual}
Selmer Bringsjord, Naveen~Sundar G., Bertram~F. Malle, and Matthias Scheutz.
\newblock Contextual deontic cognitive event calculi for ethically correct
  robots.
\newblock In \emph{International Symposium on Artificial Intelligence and
  Mathematics, {ISAIM} 2018, Fort Lauderdale, Florida, USA, January 3-5,
  2018.}, 2018.
\newblock URL
  \url{http://isaim2018.cs.virginia.edu/papers/ISAIM2018\_Ethics\_Bringsjord\_etal.pdf}.

\bibitem[Conitzer et~al.(2017)Conitzer, Sinnott-Armstrong, Borg, Deng, and
  Kramer]{Conitzer17:Moral}
Vincent Conitzer, Walter Sinnott-Armstrong, Jana~Schaich Borg, Yuan Deng, and
  Max Kramer.
\newblock Moral decision making frameworks for artificial intelligence.
\newblock In \emph{AAAI}, pages 4831--4835, 2017.
\newblock Blue Sky track.

\bibitem[Dickerson and Sandholm(2015)]{Dickerson15:FutureMatch}
John~P. Dickerson and Tuomas Sandholm.
\newblock Future{M}atch: {C}ombining human value judgments and machine learning
  to match in dynamic environments.
\newblock In \emph{AAAI}, pages 622--628, 2015.

\bibitem[Dickerson et~al.(2016)Dickerson, Manlove, Plaut, Sandholm, and
  Trimble]{Dickerson16:Position}
John~P. Dickerson, David Manlove, Benjamin Plaut, Tuomas Sandholm, and James
  Trimble.
\newblock Position-indexed formulations for kidney exchange.
\newblock In \emph{ACM EC}, 2016.

\bibitem[Dickerson et~al.(2018)Dickerson, Procaccia, and
  Sandholm]{Dickerson18:Failure}
John~P. Dickerson, Ariel~D. Procaccia, and Tuomas Sandholm.
\newblock Failure-aware kidney exchange.
\newblock \emph{Management Science}, 2018.
\newblock To appear; earlier version appeared at EC-13.

\bibitem[Elkind and Slinko(2015)]{ElkindSlinkoChapter}
Edith Elkind and Arkadii Slinko.
\newblock Rationalizations of voting rules.
\newblock In F.~Brandt, V.~Conitzer, U.~Endriss, J.~Lang, and A.~D. Procaccia,
  editors, \emph{Handbook of Computational Social Choice}. Cambridge University
  Press, 2015.
\newblock Chapter 8.

\bibitem[Ergin et~al.(2017)Ergin, S{\"o}nmez, and {\"U}nver]{Ergin17:Multi}
Haluk Ergin, Tayfun S{\"o}nmez, and M.~Utku {\"U}nver.
\newblock Multi-donor organ exchange, 2017.
\newblock Working paper.

\bibitem[Farina et~al.(2017)Farina, Dickerson, and
  Sandholm]{Farina17:Operation}
Gabriele Farina, John~P. Dickerson, and Tuomas Sandholm.
\newblock Operation frames and clubs in kidney exchange.
\newblock In \emph{IJCAI}, 2017.

\bibitem[Glorie et~al.(2014)Glorie, van~de Klundert, and
  Wagelmans]{Glorie14:Kidney}
Kristiaan Glorie, Joris van~de Klundert, and Albert Wagelmans.
\newblock Kidney exchange with long chains: An efficient pricing algorithm for
  clearing barter exchanges with branch-and-price.
\newblock \emph{Manufacturing \& Service Operations Management (MSOM)},
  16\penalty0 (4):\penalty0 498--512, 2014.

\bibitem[Greene et~al.(2016)Greene, Rossi, Tasioulas, Venable, and
  Williams]{Greene16:Embedding}
Joshua Greene, Francesca Rossi, John Tasioulas, Kristen~Brent Venable, and
  Brian~C. Williams.
\newblock Embedding ethical principles in collective decision support systems.
\newblock In \emph{AAAI}, pages 4147--4151, 2016.

\bibitem[Hajaj et~al.(2015)Hajaj, Dickerson, Hassidim, Sandholm, and
  Sarne]{Hajaj15:Strategy-Proof}
Chen Hajaj, John~P. Dickerson, Avinatan Hassidim, Tuomas Sandholm, and David
  Sarne.
\newblock Strategy-proof and efficient kidney exchange using a credit
  mechanism.
\newblock In \emph{AAAI}, pages 921--928, 2015.

\bibitem[Jia et~al.(2017)Jia, Tang, Wang, and Zhang]{Jia17:Efficient}
Zhipeng Jia, Pingzhong Tang, Ruosong Wang, and Hanrui Zhang.
\newblock Efficient near-optimal algorithms for barter exchange.
\newblock In \emph{AAMAS}, pages 362--370, 2017.

\bibitem[Kleinberg et~al.(2017)Kleinberg, Lakkaraju, Leskovec, Ludwig, and
  Mullainathan]{Kleinberg2017:Human}
Jon Kleinberg, Himabindu Lakkaraju, Jure Leskovec, Jens Ludwig, and Sendhil
  Mullainathan.
\newblock Human decisions and machine predictions.
\newblock Working Paper 23180, National Bureau of Economic Research, February
  2017.
\newblock URL \url{http://www.nber.org/papers/w23180}.

\bibitem[Li et~al.(2014)Li, Liu, Huang, and Tang]{Li14:Egalitarian}
Jian Li, Yicheng Liu, Lingxiao Huang, and Pingzhong Tang.
\newblock Egalitarian pairwise kidney exchange: Fast algorithms via linear
  programming and parametric flow.
\newblock In \emph{AAMAS}, pages 445--452, 2014.

\bibitem[Luo et~al.(2016)Luo, Tang, Wu, and Zeng]{Luo16:Approximation}
Suiqian Luo, Pingzhong Tang, Chenggang Wu, and Jianyang Zeng.
\newblock Approximation of barter exchanges with cycle length constraints.
\newblock \emph{CoRR}, abs/1605.08863, 2016.
\newblock URL \url{http://arxiv.org/abs/1605.08863}.

\bibitem[Manlove and {O'M}alley(2015)]{Manlove15:Paired}
David Manlove and Gregg {O'M}alley.
\newblock Paired and altruistic kidney donation in the {UK}: Algorithms and
  experimentation.
\newblock \emph{{ACM} Journal of Experimental Algorithmics}, 19\penalty0 (1),
  2015.

\bibitem[Mattei et~al.(2017)Mattei, Saffidine, and Walsh]{Mattei17:Mechanisms}
Nicholas Mattei, Abdallah Saffidine, and Toby Walsh.
\newblock Mechanisms for online organ matching.
\newblock In \emph{IJCAI}, 2017.

\bibitem[Montgomery et~al.(2006)Montgomery, Gentry, Marks, Warren, Hiller,
  Houp, Zachary, Melancon, Maley, Rabb, Simpkins, and
  Segev]{Montgomery06:Domino}
Robert Montgomery, Sommer Gentry, William~H Marks, Daniel~S Warren, Janet
  Hiller, Julie Houp, Andrea~A Zachary, J~Keith Melancon, Warren~R Maley, Hamid
  Rabb, Christopher Simpkins, and Dorry~L Segev.
\newblock Domino paired kidney donation: a strategy to make best use of live
  non-directed donation.
\newblock \emph{The Lancet}, 368\penalty0 (9533):\penalty0 419--421, 2006.

\bibitem[Noothigattu et~al.(2018)Noothigattu, Gaikwad, Awad, Dsouza, Rahwan,
  Ravikumar, and Procaccia]{Noothigattu18:Voting}
Ritesh Noothigattu, Snehalkumar (Neil)~S. Gaikwad, Edmond Awad, Sohan Dsouza,
  Iyad Rahwan, Pradeep Ravikumar, and Ariel~D. Procaccia.
\newblock A voting-based system for ethical decision making.
\newblock In \emph{{AAAI}}, pages 1587--1594. {AAAI} Press, 2018.

\bibitem[O'Neil(2017)]{ONeil17:Weapons}
Cathy O'Neil.
\newblock \emph{Weapons of math destruction: How big data increases inequality
  and threatens democracy}.
\newblock Broadway Books, 2017.

\bibitem[Pereira and Saptawijaya(2017)]{Pereira17:Agent}
Lu\'is~Moniz Pereira and Ari Saptawijaya.
\newblock Agent morality via counterfactuals in logic programming.
\newblock In \emph{Proceedings of the Bridging@CogSci Workshop}, pages 39--53,
  London, UK, 2017.

\bibitem[Rahwan and Simari(2009)]{Rahwan09:Argumentation}
Iyad Rahwan and Guillermo~R Simari.
\newblock \emph{Argumentation in artificial intelligence}, volume~47.
\newblock Springer, 2009.

\bibitem[Rees et~al.(2009)Rees, Kopke, Pelletier, Segev, Rutter, Fabrega,
  Rogers, Pankewycz, Hiller, Roth, Sandholm, {\"{U}}nver, and
  Montgomery]{Rees09:Nonsimultaneous}
Michael Rees, Jonathan Kopke, Ronald Pelletier, Dorry Segev, Matthew Rutter,
  Alfredo Fabrega, Jeffrey Rogers, Oleh Pankewycz, Janet Hiller, Alvin Roth,
  Tuomas Sandholm, Utku {\"{U}}nver, and Robert Montgomery.
\newblock A nonsimultaneous, extended, altruistic-donor chain.
\newblock \emph{New England Journal of Medicine}, 360\penalty0 (11):\penalty0
  1096--1101, 2009.

\bibitem[Roth et~al.(2005{\natexlab{a}})Roth, S{\"{o}}nmez, and
  {\"U}nver]{Roth05:Kidney}
Alvin Roth, Tayfun S{\"{o}}nmez, and Utku {\"U}nver.
\newblock A kidney exchange clearinghouse in {N}ew {E}ngland.
\newblock \emph{American Economic Review}, 95\penalty0 (2):\penalty0 376--380,
  2005{\natexlab{a}}.

\bibitem[Roth et~al.(2005{\natexlab{b}})Roth, S{\"{o}}nmez, and
  {\"U}nver]{Roth05:Pairwise}
Alvin Roth, Tayfun S{\"{o}}nmez, and Utku {\"U}nver.
\newblock Pairwise kidney exchange.
\newblock \emph{Journal of Economic Theory}, 125\penalty0 (2):\penalty0
  151--188, 2005{\natexlab{b}}.

\bibitem[Roth et~al.(2004)Roth, Sonmez, and Unver]{Roth04:Kidney}
Alvin~E. Roth, Tayfun Sonmez, and M.~Utku Unver.
\newblock Kidney exchange.
\newblock \emph{Quarterly Journal of Economics}, 119\penalty0 (2):\penalty0
  457--488, 2004.

\bibitem[Tolchinsky et~al.(2012)Tolchinsky, Modgil, Atkinson, McBurney, and
  Cort{\'e}s]{Tolchinsky12:Deliberation}
Pancho Tolchinsky, Sanjay Modgil, Katie Atkinson, Peter McBurney, and Ulises
  Cort{\'e}s.
\newblock Deliberation dialogues for reasoning about safety critical actions.
\newblock \emph{Autonomous Agents and Multi-Agent Systems}, 25\penalty0
  (2):\penalty0 209, 2012.

\bibitem[Toulis and Parkes(2015)]{Toulis15:Design}
Panos Toulis and David~C. Parkes.
\newblock Design and analysis of multi-hospital kidney exchange mechanisms
  using random graphs.
\newblock \emph{Games and Economic Behavior}, 91:\penalty0 360--382, 2015.

\bibitem[Turing(1950)]{Turing50:Computing}
A.~M. Turing.
\newblock Computing machinery and intelligence.
\newblock \emph{Mind}, 59\penalty0 (236):\penalty0 433--460, 1950.
\newblock ISSN 00264423.

\bibitem[{UNOS}(2015)]{UNOS15:Revising}
{UNOS}.
\newblock Revising kidney paired donation pilot program priority points, 2015.
\newblock OPTN/UNOS Public Comment Proposal.

\bibitem[Wallach and Allen(2008)]{Wallach08:Moral}
Wendell Wallach and Colin Allen.
\newblock \emph{Moral Machines: Teaching Robots Right from Wrong}.
\newblock {Oxford University Press}, 2008.

\bibitem[Wallach and Allen(2010)]{Wallach10:Moral}
Wendell Wallach and Colin Allen.
\newblock \emph{Moral Machines: Teaching Robots Right from Wrong}, chapter~6,
  pages 83--98.
\newblock Oxford University Press, Inc., New York, NY, USA, 2010.
\newblock ISBN 0199737975, 9780199737970.

\end{thebibliography}

\end{document}